\documentclass[conference]{IEEEtran}
\IEEEoverridecommandlockouts

\usepackage{cite}
\usepackage{amsmath,amssymb,amsfonts}
\usepackage{algorithmic}
\usepackage{graphicx}
\usepackage{textcomp}
\usepackage{xcolor}
\def\BibTeX{{\rm B\kern-.05em{\sc i\kern-.025em b}\kern-.08em
    T\kern-.1667em\lower.7ex\hbox{E}\kern-.125emX}}
\begin{document}

\title{Hyperbolic Structured Classification for Robust Single Positive Multi-label Learning}

\author{\IEEEauthorblockN{Yiming Lin}
\IEEEauthorblockA{\textit{School of Advanced Technology} \\
\textit{Xi'an Jiaotong-Liverpool University}\\
Suzhou, China \\
yiming.lin21@student.xjtlu.edu.cn}
\and
\IEEEauthorblockN{Shang Wang}
\IEEEauthorblockA{\textit{School of Advanced Technology} \\
\textit{Xi'an Jiaotong-Liverpool University}\\
Suzhou, China \\
shang.wang24@student.xjtlu.edu.cn}
\and
\IEEEauthorblockN{Junkai Zhou}
\IEEEauthorblockA{\textit{School of Data Science} \\
\textit{Chinese University of Hong Kong, Shenzhen}\\
Shenzhen, China \\
122090818@link.cuhk.edu.cn}
\and
\IEEEauthorblockN{Qiufeng Wang}
\IEEEauthorblockA{\textit{School of Advanced Technology} \\
\textit{Xi'an Jiaotong-Liverpool University}\\
Suzhou, China \\
qiufeng.wang@xjtlu.edu.cn}
\and
\IEEEauthorblockN{Xiao-Bo Jin}
\IEEEauthorblockA{\textit{School of Advanced Technology} \\
\textit{Xi'an Jiaotong-Liverpool University}\\
Suzhou, China \\
xiaobo.jin@xjtlu.edu.cn}
\and
\IEEEauthorblockN{Kaizhu Huang}
\IEEEauthorblockA{\textit{Digital Innovation Research Center} \\
\textit{Duke Kunshan University}\\
Suzhou, China \\
kaizhu.huang@dukekunshan.edu.cn}
}

\maketitle

\begin{abstract}
Single Positive Multi-Label Learning (SPMLL) addresses the challenging scenario where each training sample is annotated with only one positive label despite potentially belonging to multiple categories, making it difficult to capture complex label relationships and hierarchical structures. While existing methods implicitly model label relationships through distance-based similarity, lacking explicit geometric definitions for different relationship types.

To address these limitations, we propose the first hyperbolic classification framework for SPMLL that represents each label as a hyperbolic ball rather than a point or vector, enabling rich inter-label relationship modeling through geometric ball interactions. Our ball-based approach naturally captures multiple relationship types simultaneously: inclusion for hierarchical structures, overlap for co-occurrence patterns, and separation for semantic independence. Further, we introduce two key component innovations: a temperature-adaptive hyperbolic ball classifier and a physics-inspired double-well regularization that guides balls toward meaningful configurations.

To validate our approach, extensive experiments on four benchmark datasets (MS-COCO, PASCAL VOC, NUS-WIDE, CUB-200-2011) demonstrate competitive performance with superior interpretability compared to existing methods. Furthermore, statistical analysis reveals strong correlation between learned embeddings and real-world co-occurrence patterns, establishing hyperbolic geometry as a more robust paradigm for structured classification under incomplete supervision.
    

\end{abstract}
\begin{IEEEkeywords}
Hyperbolic geometry, Multi-label learning, Single positive multi-lable learning
\end{IEEEkeywords}

\section{Introduction}


Multi-label recognition (MLR) is a fundamental problem in domains such as computer vision, natural language processing, multimedia retrieval, and medical diagnosis. In practical scenarios, objects and concepts often possess multiple semantic labels simultaneously—for example, an image may contain both "person" and "bicycle", while a document might be tagged with "machine learning", "computer vision", and "artificial intelligence." However, acquiring complete multi-label annotations is prohibitively expensive, requiring substantial human effort, domain expertise, and time. As a result, some datasets may contain only one positive label per sample, even though the instances may belong to multiple categories. This incomplete supervision makes it particularly challenging to naturally model complex inter-category relationships that are essential for accurate multi-label recognition. The widespread availability of such single-positive data motivates the development of learning strategies that can effectively leverage incomplete supervision while naturally expressing semantic dependencies among labels.

Single Positive Multi-Label Learning (SPMLL)~\cite{cole2021multi} formalizes this weakly supervised setting, where each training instance is annotated with only one positive label while all other potentially relevant labels remain unobserved. This setting introduces fundamental challenges that go beyond those in conventional multi-label learning. The uncertainty of missing labels—where unobserved labels may correspond to either true negatives or unlabeled positives—leads to ambiguous and noisy supervision. Moreover, the supervision signal is extremely sparse, resulting in severe class imbalance that distorts learning dynamics and biases model predictions. Compounding these issues is the lack of label co-occurrence information, which prevents models from leveraging rich dependencies among labels that are otherwise critical for accurate multi-label recognition.

While recent work has made progress in SPMLL~\cite{lv2019weakly,cole2021multi,zhang2021simple,kim2022large,liu2023revisiting}, most existing methods~\cite{ding2023exploring,wang2023hierarchical} remain fundamentally constrained by their reliance on Euclidean geometry and explicit graph-based modeling of label relationships. Typically, these approaches use graph neural networks with predefined similarity matrices to encode label dependencies, often requiring external knowledge or auxiliary annotations to construct meaningful graphs. However, such models are primarily designed to capture pairwise semantic similarity, overlooking richer structural relations such as hierarchical inclusion, mutual exclusivity, and partial overlap between labels. Furthermore, Euclidean space itself is inherently limited in expressiveness when modeling the hierarchical structures that are prevalent in multi-label taxonomies. This geometric mismatch between model assumptions and the underlying data manifold ultimately restricts both the capacity of learned representations and the interpretability of the resulting classification decisions.

Hyperbolic geometry provides a compelling alternative through its unique capacity for naturally and automatically representing complex relational structures via interpretable ball-based embeddings~\cite{NEURIPS2022_d51ab0fc}. Unlike Euclidean space where labels are typically represented as points with uniform distance relationships, hyperbolic space enables each label to be represented as a ball with varying size and position, creating naturally interpretable geometric interactions. The negative curvature of hyperbolic space naturally and automatically supports multiple types of ball-to-ball relationships: ball inclusion for hierarchical structures (parent categories encompassing child categories), ball overlap for co-occurrence patterns (frequently co-appearing labels sharing regions), and ball separation for semantic independence (unrelated concepts maintaining distinct boundaries). This geometric richness allows a single unified framework to simultaneously and naturally capture hierarchical, co-occurrence, and similarity relationships that are essential for multi-label learning, without requiring explicit relationship graphs, external supervision, or sacrificing interpretability for performance.

Building on these insights, we propose a novel hyperbolic structured classification framework for SPMLL that operates within hyperbolic space. Our approach establishes an end-to-end pipeline that projects CLIP-derived image features into the Poincaré ball model, enabling the model to capture hierarchical label correlation in a geometry-aware manner with rich interpretability of the learned relationships. This unified framework jointly optimizes feature projections, label representations, and confidence scaling in hyperbolic space, allowing rich structural information to emerge naturally during training while providing transparent geometric visualization of inter-category dependencies. Unlike existing methods that sacrifice interpretability for performance, our framework maintains both competitive classification accuracy and superior interpretability through its naturally interpretable ball-based geometric interactions.

At the heart of our framework is a temperature-adaptive hyperbolic ball classifier, which represents each label as a hyperbolic ball in the Poincaré model. Each ball is equipped with a learnable, class-specific temperature parameter that flexibly adjusts the curvature-aware decision boundaries. To provide smoother regularization in hyperbolic space, we introduce a physics-inspired double-well loss function that defines a bistable energy landscape, encouraging confident separation between positive and negative examples. These components are jointly trained through a unified multi-objective optimization strategy that combines binary cross-entropy, double-well regularization, and a uniformity loss. Together, they enable the model to learn representations that are not only discriminative but also geometrically structured.

We evaluate our method on four benchmark datasets MS-COCO~\cite{lin2014microsoft}, PASCAL VOC~\cite{hoiem2009pascal}, NUS-WIDE~\cite{chua2009nus} and CUB-200-2011~\cite{wah2011caltech}, demonstrating competitive performance while providing superior interpretability compared to existing SPMLL approaches. Ablation studies confirm the effectiveness of each core component, and visualization analyses reveal that the learned hyperbolic embeddings align closely with underlying semantic label hierarchies, providing transparent insights into discovered label relationships. Statistical analysis validates that our naturally learned geometric relationships automatically capture real-world co-occurrence patterns, establishing a new paradigm where strong performance and clear explainability are achieved simultaneously. To the best of our knowledge, this work constitutes the first successful application of hyperbolic geometry to single positive multi-label learning, offering a new paradigm for interpretable structured classification under incomplete supervision.

%

\section{Literature Review}

\subsection{Classification in Hyperbolic Space}

Traditional approaches to hierarchical multi-label classification are often constrained by the limited expressive power of Euclidean space. While some methods use distance-based similarity to model label relationships implicitly, they typically lack explicit geometric formulations for diverse relational types. To overcome these limitations, recent studies have explored hyperbolic geometry as a natural fit for hierarchy modeling. For example, HyperIM jointly embeds documents and labels into hyperbolic space, explicitly capturing semantic dependencies between words and labels to enhance classification performance~\cite{chen2020hyperbolic}.

To move beyond manually defined hierarchies, some approaches jointly learn label embeddings and classifier parameters, enabling models to discover latent structures that better reflect the data manifold~\cite{chatterjee2021joint}. Moreover, several alternative geometric frameworks have been proposed. Lorentzian embeddings offer a numerically stable alternative to Poincaré models for large-scale applications~\cite{nickel2018learning}, while entailment cones encode partial orders through directional constraints in hyperbolic space~\cite{ganea2018hyperbolic}. However, many methods still rely on Euclidean-based classifiers or losses, creating a geometric mismatch that limits performance.

In response to this geometric mismatch, HypEmo fully embeds the entire classification pipeline within the Poincaré model and employs a hyperbolic distance-weighted loss, significantly improving performance in fine-grained sentiment classification~\cite{chen2023label}. Hyperbolic SVM further extends support vector machines into hyperbolic geometry, enabling more structurally consistent decision surfaces and yielding notable gains in hierarchical settings~\cite{cho2019large}.

Nonetheless, most existing hyperbolic methods continue to represent labels as points or vectors, lacking the capacity to explicitly model diverse relational semantics—such as hierarchy, overlap, and independence—within a unified geometric framework. This limitation reduces their effectiveness in structured classification tasks under weak supervision or partial labeling.

\subsection{Single Positive Multi-Label Learning}

Single Positive Multi-Label Learning (SPMLL) addresses the challenge of learning from training data where each instance is annotated with only one positive label, while all others remain unobserved—potentially including false negatives. This problem was first formalized by Cole et al.~\cite{cole2021multi}, who proposed several loss-based strategies to mitigate missing supervision, including weak negative sampling, label smoothing, and online label estimation~\cite{zhou2022acknowledging}. While effective to some extent, these methods struggle with extreme label sparsity and inherent uncertainty.

To recover missing labels, pseudo-labeling has become a dominant strategy. Early efforts such as SPLC generated labels from high-confidence predictions~\cite{xu2022one}, while co-pseudo training frameworks introduced adaptive thresholds and sample selection mechanisms to combat confirmation bias and class imbalance~\cite{hu2025co}. Recent improvements include the use of pre-trained vision-language models like CLIP to generate more reliable positive pseudo-labels under weak supervision~\cite{xing2024vision}. Other models, such as SMILE, extend this direction by applying label enhancement through unbiased risk estimation, thereby expanding supervision from single positive labels to more comprehensive label structures~\cite{xu2022one}. However, these strategies often rely on pairwise similarity, failing to capture richer, structured semantics.

To address this, several approaches explicitly model inter-label relationships to guide reasoning about unobserved labels. SigRL leverages graph-based reasoning and semantic feature reconstruction to encode label dependencies~\cite{zhang2025semantic}, while SpliceMix reduces contextual co-occurrence bias through cross-scale and semantic blending augmentation~\cite{wang2025splicemix}. These methods offer greater structure-awareness, but typically operate in Euclidean space, which limits their expressiveness when modeling hierarchical, overlapping, or disjoint semantics.

In parallel, robust loss functions have been developed to handle false negatives and supervision noise. IGNORE introduces an information-gap-based criterion to reject unreliable negatives~\cite{song2024ignore}, and GRLoss combines pseudo-labels with soft risk estimation for better flexibility~\cite{chen2024boosting}. Other strategies attempt to decouple overly strong negative assumptions—for example, by pushing only one label pair apart at a time~\cite{ding2023exploring} or prompting semantic alignment through CLIP-based cues. Despite these advances, few methods directly tackle the geometric inconsistency between the structure of label relationships and the underlying decision space.

\section{Method}

\subsection{Hyperbolic Geometry Foundations}

We work in the Poincaré ball model $\mathbb{D}^n = \{x \in \mathbb{R}^n : \|x\|_2 < 1\}$ equipped with the Riemannian metric $g^D_x = \lambda^2_x g^E$, where $\lambda_x = \frac{2}{1-\|x\|^2}$ is the conformal factor. The key advantage of hyperbolic geometry lies in its exponential volume growth: regions near the origin naturally accommodate broad correlation patterns with extensive influence, while areas toward the boundary are suited for specific correlation patterns requiring fine-grained distinctions.

The fundamental distance metric in hyperbolic space measures the shortest path between any two points:
\begin{equation}
d_{\mathbb{D}}(u, v) = \text{arccosh}\left(1 + 2\frac{\|u - v\|^2}{(1 - \|u\|^2)(1 - \|v\|^2)}\right) \label{eq:distance}
\end{equation}

Möbius addition provides the hyperbolic analogue of vector addition, enabling meaningful combination of hyperbolic vectors:
\begin{equation}
x \oplus y = \frac{(1 + 2\langle x, y \rangle + \|y\|^2)x + (1 - \|x\|^2)y}{1 + 2\langle x, y \rangle + \|x\|^2\|y\|^2} \label{eq:mobius}
\end{equation}

The exponential map at the origin projects tangent vectors onto the hyperbolic manifold:
\begin{equation}
\exp_0(v) = \tanh(\|v\|) \frac{v}{\|v\|} \label{eq:exp}
\end{equation}

These operations form the foundation for Möbius linear transformations that map Euclidean features to hyperbolic space while preserving manifold structure, enabling us to build label correlation learning frameworks that respect the geometric properties of hyperbolic space.
\subsection{Problem Formulation and Label Correlation Modeling}

SPMLL provides only one positive label per sample while leaving other relevant labels unobserved, making it challenging to directly observe the full spectrum of label correlation patterns in the data. Real-world label correlations exhibit diverse statistical patterns: some labels frequently co-occur, others exhibit hierarchical statistical relationships, while many remain statistically independent. This incomplete observation creates fundamental challenges for learning comprehensive correlation structures from minimal supervision.

\textbf{Limitations of Traditional Approaches.} Traditional methods model labels as points in Euclidean space, relying on uniform distance to encode correlations. This approach lacks expressiveness, capturing only binary relationships and failing to represent nuanced correlations from real-world data. The rigidity of Euclidean geometry also hinders the modeling of asymmetric relations like containment and exclusion, particularly within hierarchical structures. Since Euclidean distance is symmetric, it cannot naturally represent directional dependencies such as one label implying another.

\textbf{Hyperbolic Ball-based Solution.} We propose representing each label as a hyperbolic ball rather than a point, enabling automatic encoding of multiple correlation types discovered from data statistics. Each label embedding $c_i \in \mathbb{D}^n$ defines a hyperbolic ball with radius $r_i = \frac{1 - \|c_i\|^2}{2\|c_i\|}$ and center $c_i^*$ positioned along the ray from origin through $c_i$. As embeddings move toward the boundary, radius increases, creating natural coverage gradients where different radial positions accommodate different correlation pattern complexities—broad correlations near origin, specific correlations near boundary.

\textbf{Emergent Correlation Encoding.} Ball interactions automatically encode three fundamental label correlation patterns discovered from training data (Figure~\ref{fig:hyperbolic_relationships}): \textit{(1) Hierarchical Correlations} through ball containment ($B_{\text{broader}} \supseteq B_{\text{specific}}$) reflecting statistical subsumption relationships, \textit{(2) Co-occurrence Correlations} through overlapping regions capturing frequent co-appearance patterns in the dataset, and \textit{(3) Independence Patterns} through separated regions reflecting statistical non-dependence between label pairs. These correlation patterns emerge organically during training without any predefined structure, creating a self-organizing system where ball configurations automatically adapt to reflect the underlying statistical correlation structure discovered from SPMLL data. The next step is to design the mathematical framework for implementing these hyperbolic ball representations. The next step is to design the mathematical framework for implementing these hyperbolic ball representations.

\subsection{Hyperbolic Ball-based Framework Design}

\textbf{Mathematical Foundation.} Each label embedding $c_i \in \mathbb{D}^n$ with norm $\rho_i = \|c_i\|$ defines a hyperbolic ball:
\begin{align}
r_i &= \frac{1 - \rho_i^2}{2\rho_i} \quad \text{(radius)} \label{eq:radius}\\
c_i^* &= c_i \left(1 + \frac{r_i}{\rho_i}\right) \quad \text{(center)} \label{eq:center}
\end{align}
This geometric construction ensures that balls near the origin have smaller radii (suitable for broad correlation patterns) while those near the boundary have larger radii (suitable for specific correlation patterns), naturally organizing correlation complexity across hyperbolic space according to the statistical structure in the data.

\begin{figure*}[t]
\centering
\includegraphics[width=1.4\columnwidth]{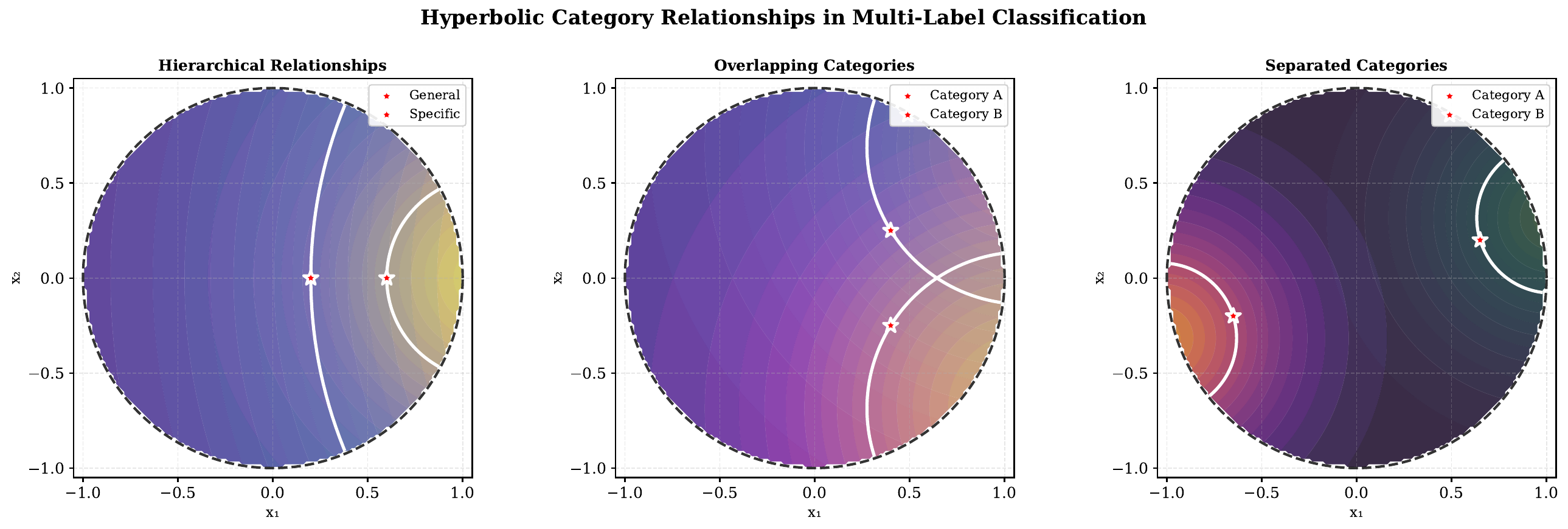}
\caption{Label correlation encoding through hyperbolic ball interactions. \textbf{Left:} Hierarchical correlations with nested response patterns discovered from data. \textbf{Center:} Co-occurrence correlations through intersecting decision regions. \textbf{Right:} Statistical independence via separated geometric regions. Heat maps show response intensities, white contours indicate decision boundaries (0.5), and red stars mark label centers. These patterns emerge automatically from learning statistical correlation structures in the training data.}
\label{fig:hyperbolic_relationships}
\end{figure*}

\textbf{Classification Mechanism.} For input embedding $x \in \mathbb{D}^n$, membership in label $l_i$ is determined by the signed distance from the input to the ball boundary:
\begin{equation}
\text{membership}(x, l_i) = r_i - \|c_i^* - x\| \label{eq:membership}
\end{equation}
where $c_i^*$ is the ball center computed as in Eq.~(\ref{eq:center}). Positive values indicate membership (inside the ball), negative values indicate non-membership. Figure~\ref{fig:hyperbolic_relationships} visualizes how different ball configurations automatically encode distinct correlation types through classifier response patterns.

\textbf{Geometric Adaptivity Challenge.} While ball membership provides the foundation for classification, the non-uniform curvature of hyperbolic space poses a significant challenge: classification mechanisms must adapt to different radial positions to effectively model varying correlation complexities. This necessitates the development of position-adaptive mechanisms and specialized regularization techniques.

\subsection{Adaptive Classification and Regularization Framework}

\textbf{Non-Uniform Curvature Challenge.} Hyperbolic space's non-uniform curvature creates distinct challenges for modeling label correlations at different geometric positions: regions near the origin are relatively flat and naturally accommodate broad correlation patterns, while boundary areas experience extreme curvature compression and are suited for fine-grained correlation distinctions. This geometric heterogeneity causes labels at different radial positions to require fundamentally different classification mechanisms—broad correlation regions near the origin need smooth decision boundaries, while specific correlation patterns near the boundary demand precise discrimination.

We introduce position-dependent scaling with the classification score:
\begin{equation}
s_i(x) = \frac{\alpha_i}{\tau} \cdot (r_i - \|c_i^* - x\|) \label{eq:score}
\end{equation}
where $\alpha_i = \frac{2}{1 - \rho_i^2}$ is the hyperbolic scaling factor, $\tau$ is a learnable temperature parameter, and $c_i^*$ is the ball center from Eq.~(\ref{eq:center}). As labels approach the boundary ($\rho_i \to 1$), $\alpha_i$ grows exponentially, amplifying small geometric differences in high-curvature regions.

\textbf{Temperature Adaptation.} Temperature $\tau$ provides adaptive calibration across different correlation pattern complexities at various radial positions. Figure~\ref{fig:temperature_analysis} shows temperature scaling creates position-aware boundaries that adapt to local correlation structures: low temperatures ($\tau = 0.1$) produce sharp boundaries suitable for distinct correlation patterns, higher temperatures ($\tau = 5.0$) create smooth surfaces appropriate for overlapping correlation regions. Joint optimization automatically discovers optimal temperature scales for different correlation pattern types across the label space, crucial for learning robust correlations from SPMLL's incomplete supervision.

\textbf{Synergistic Adaptivity.} Temperature scaling and double-well regularization work synergistically: the former adapts decision boundaries to local correlation complexities, while the latter guides embeddings to geometrically favorable regions. This dual adaptivity enables comprehensive correlation discovery from minimal supervision across the hyperbolic space.

\subsection{Training Objective and Loss Functions}

Having established the adaptive classification framework, we now present the comprehensive training objective that integrates these components with specialized loss functions to guide the automatic discovery of meaningful correlation structures.

Our training framework integrates multiple complementary loss terms that address different aspects of label correlation learning in hyperbolic space for SPMLL. The complete objective balances classification accuracy with geometric regularization to ensure stable correlation discovery and meaningful spatial organization that reflects the underlying statistical structure:

\begin{equation}
\mathcal{L} = \mathcal{L}_{\text{cls}} + \lambda_1 \mathcal{L}_{\text{reg}} + \lambda_2 \mathcal{L}_{\text{uni}} \label{eq:total_loss}
\end{equation}

\textbf{Classification Loss Component.} The primary supervision term $\mathcal{L}_{\text{cls}}$ applies standard binary cross-entropy to our position-adaptive classification scores, providing direct supervision for learning label correlations from the observed positive labels. Input images are first processed through a pre-trained CLIP encoder to obtain Euclidean visual features, which are then mapped to hyperbolic space via learnable Möbius linear transformations that preserve the manifold structure while enabling the discovery of label correlation patterns specific to each dataset.

\textbf{Double-Well Regularization.} Our physics-inspired regularization $\mathcal{L}_{\text{reg}}$ creates a bistable energy landscape to guide label embedding organization:

\begin{align}
\mathcal{L}_{\text{reg}} = \sum_{i=1}^K \Big[-&\exp(-\beta_1(\rho_i - c_1)^2)(1-\exp(-\beta_2\rho_i^2)) \nonumber \\
&- \exp(-\beta_1(\rho_i - c_2)^2)(1-\exp(-\beta_2(1-\rho_i)^2))\Big] \label{eq:reg_loss}
\end{align}

This creates two attractive wells at $c_1 \approx 0.1$ and $c_2 \approx 0.9$ for organizing correlation patterns by complexity, with detailed analysis provided in Section~3.7.

\textbf{Uniformity Constraint.} The uniformity loss $\mathcal{L}_{\text{uni}} = \frac{1}{K(K-1)} \sum_{i \neq j} |\langle \hat{c}_i, \hat{c}_j \rangle|$ prevents correlation pattern collapse through angular diversity, where $\hat{c}_i$ denotes the normalized embedding. Combined with Riemannian optimization that ensures manifold-respecting updates, the complete framework enables automatic discovery of diverse label correlation patterns from minimal SPMLL supervision.

\subsection{Temperature Adaptation Analysis}

The temperature adaptation mechanism addresses the fundamental challenge of achieving consistent classification performance across hyperbolic space's non-uniform curvature. Different radial positions in the Poincaré disk exhibit vastly different geometric properties: near-origin regions are relatively flat and accommodate broad correlation patterns, while boundary-proximate areas experience extreme curvature compression suitable for fine-grained correlation distinctions.

\begin{figure}[htbp]
\centering
\includegraphics[width=\linewidth]{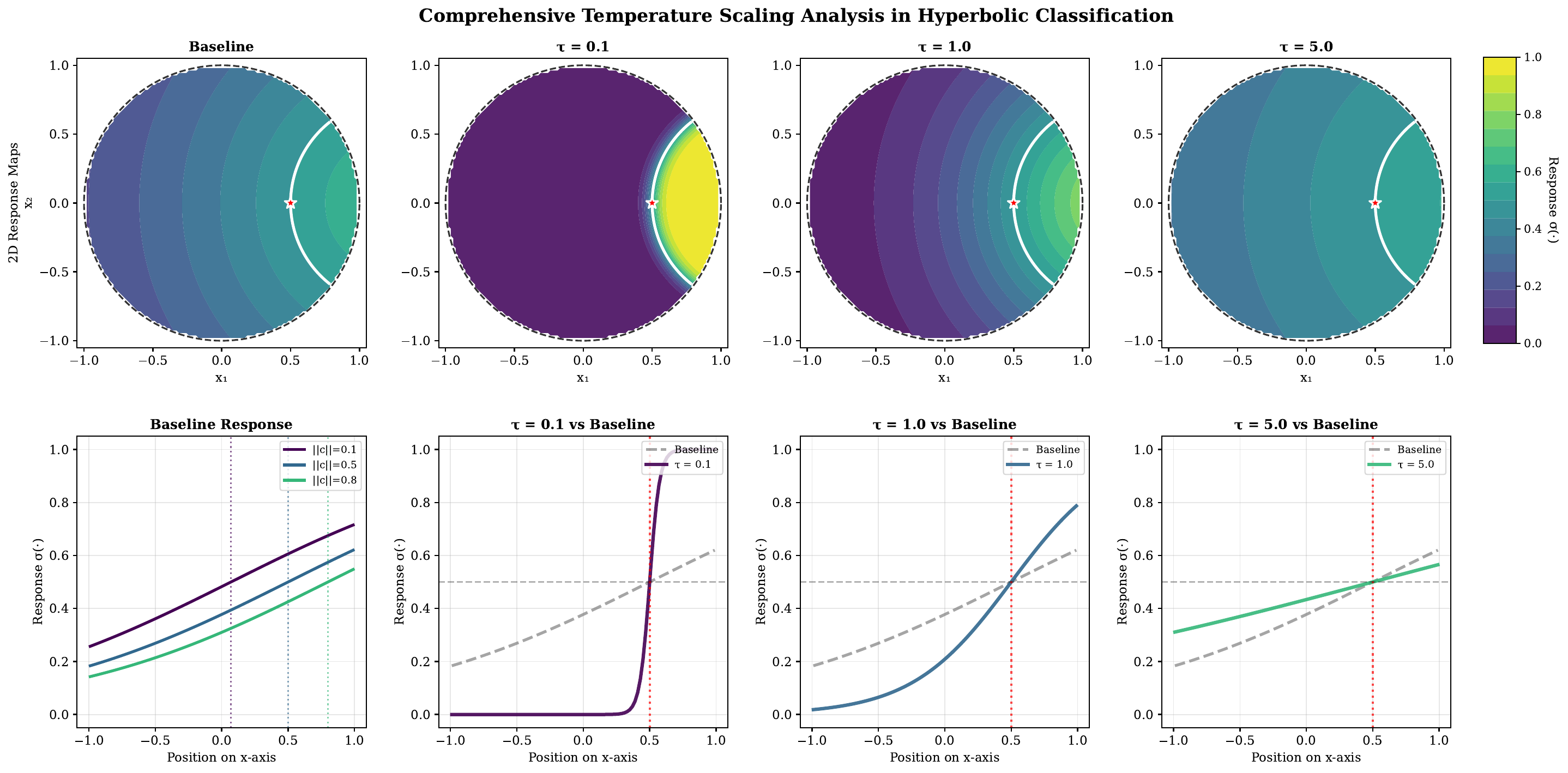}
\caption{Comprehensive temperature scaling analysis combining baseline and temperature-adaptive methods. Top row displays 2D response maps for baseline and three temperature settings ($\tau = 0.1, 1.0, 5.0$) applied to mid-position label ($\|c\| = 0.5$). Bottom row shows 1D response curves: (left) baseline response across different class positions, (right three) direct comparisons between baseline and temperature-scaled methods. This unified analysis demonstrates the consistent superiority of temperature scaling across all geometric positions in hyperbolic space.}
\label{fig:temperature_analysis}
\end{figure}

\textbf{Position-Dependent Response Characteristics.} Figure~\ref{fig:temperature_analysis} demonstrates how temperature scaling creates adaptive decision boundaries across representative geometric positions. The visualization reveals that without temperature adaptation, labels at different radial positions exhibit inconsistent response distributions—a critical issue for learning reliable correlations. The temperature mechanism provides position-aware calibration that ensures consistent correlation learning across all geometric locations.

\textbf{Adaptive Boundary Formation.} The learnable temperature parameter $\tau$ automatically adapts to the local geometric context of each correlation pattern. Low temperatures create sharp decision boundaries suitable for well-separated correlation patterns, while higher temperatures generate smooth boundaries appropriate for overlapping correlation regions. This adaptive mechanism enables the framework to automatically discover optimal decision boundaries for different types of correlation patterns without manual tuning.

\subsection{Double-Well Regularization Analysis}

\begin{figure}[!t]
\centering
\includegraphics[width=\linewidth]{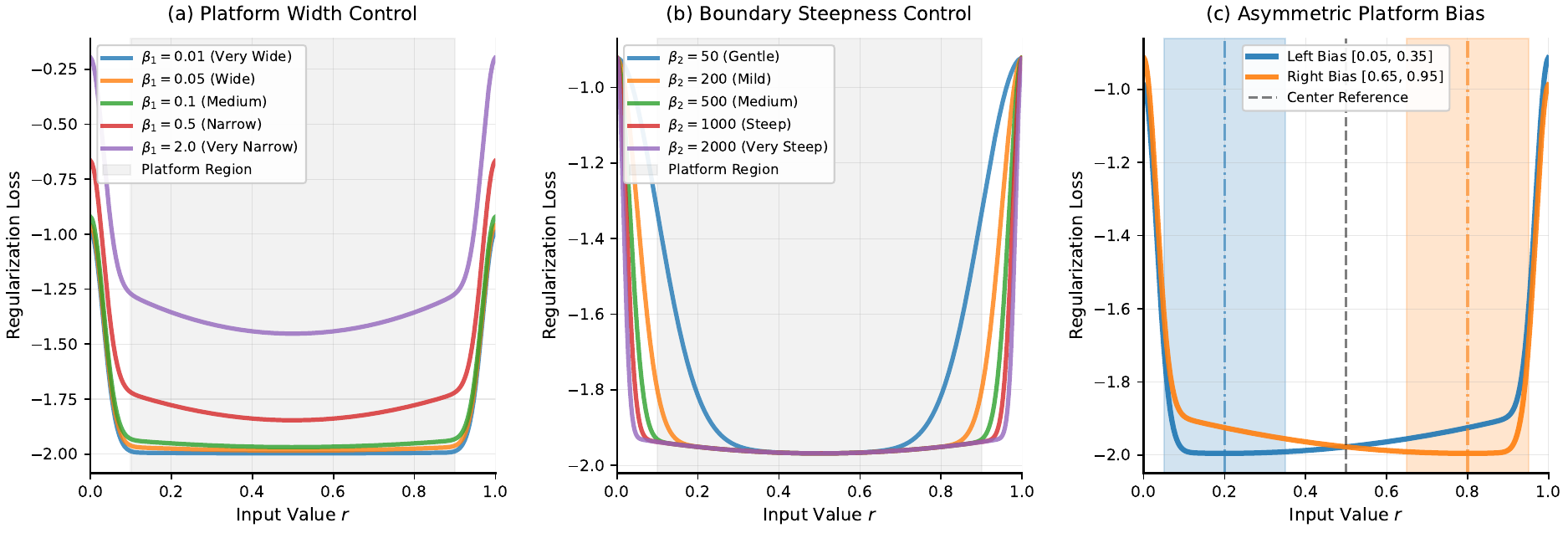}
\caption{Double-well regularizer parameter analysis. \textbf{(a)} Platform width control via $\beta_1$ balances correlation pattern stability. \textbf{(b)} Boundary steepness control via $\beta_2$ enforces geometric constraints. \textbf{(c)} Asymmetric platform configuration adapts to correlation complexity distributions.}
\label{fig:double_well_analysis}
\end{figure}

The double-well regularizer complements temperature adaptation by providing spatial organization guidance for correlation pattern learning. As illustrated in Figure~\ref{fig:double_well_analysis}, the energy landscape creates two attractive regions that naturally separate correlation patterns by complexity.

\textbf{Energy Landscape Design.} The regularizer is controlled by four key parameters that shape the correlation-organizing energy landscape. Platform width $\beta_1 \in [0.05, 0.1]$ balances embedding stability and precision for correlation learning, while boundary steepness $\beta_2 \in [500, 1000]$ enforces geometric constraints without causing gradient instability. The asymmetric platform configuration enables flexible adaptation to different correlation pattern distributions in various datasets.

\section{Experiments}

We conduct comprehensive experiments to validate the effectiveness of our hyperbolic structured classification framework for single positive multi-label learning. This section presents detailed experimental setup, comparisons with state-of-the-art methods, ablation studies, and in-depth analysis of the learned hyperbolic representations.

\subsection{Experimental Setup}

\textbf{Datasets.} We evaluate our method on four widely-used benchmark datasets for multi-label learning: MS-COCO~\cite{lin2014microsoft}, PASCAL VOC 2007~\cite{hoiem2009pascal}, NUS-WIDE~\cite{chua2009nus}, and CUB-200-2011~\cite{wah2011caltech}. Following the single positive multi-label learning protocol, each training image is annotated with only one positive label while all other labels remain unobserved. MS-COCO contains 80 object categories with approximately 80K training images. PASCAL VOC 2007 includes 20 object classes with around 5K training images. NUS-WIDE consists of 81 concept categories with 269K images. CUB-200-2011 provides 200 fine-grained bird species with clear hierarchical relationships, making it particularly suitable for evaluating hyperbolic geometry's ability to capture semantic hierarchies.

\textbf{Implementation Details.} We implement our method using PyTorch~\cite{imambi2021pytorch} with geoopt library~\cite{kochurov2020geoopt} for Riemannian optimization and pre-trained CLIP RN50 as visual backbone. Hyperbolic operations use Poincaré ball model with RiemannianAdam (lr=$1 \times 10^{-4}$) for geometric parameters and Adam (lr=$1 \times 10^{-5}$) for image encoder. Loss weights: $\lambda_1=10.0$ (double-well), $\lambda_2=1.0$ (uniformity). Training: 60 epochs, batch size 128, gradient clipping norm 1.0.

\textbf{Evaluation Metrics.} We adopt mean Average Precision (mAP) as the primary evaluation metric, following standard practice in multi-label learning. For geometric analysis, we provide visualizations of learned hyperbolic embeddings and statistical analysis of temperature parameter distributions.

\subsection{Comparison with State-of-the-art Methods}

We compare our method against several representative approaches for single positive multi-label learning, including SPLC, SCPNet, and other recent methods. All baseline methods use the same data preprocessing and evaluation protocols to ensure fair comparison.

\textbf{Main Results.} Table~\ref{tab:main_results} presents the comprehensive comparison on four benchmark datasets. Our hyperbolic structured classification framework achieves competitive performance while providing superior interpretability compared to existing methods. Although SCPNet achieves higher accuracy on some datasets, our method offers unique geometric interpretability that existing methods lack. Notably, we achieve improvements over traditional methods (SPLC, Hill, etc.) on most datasets, particularly on NUS-WIDE and CUB-200-2011, validating the potential of hyperbolic geometry for modeling hierarchical label relationships under incomplete supervision while maintaining transparent geometric insights.

\begin{table}[htbp]
\centering
\caption{Comparison with state-of-the-art methods on four benchmark datasets. All results are reported as mAP (\%).}
\label{tab:main_results}
\begin{tabular}{|l|c|c|c|c|c|}
\hline
Method & COCO & VOC & NUS-WIDE & CUB & Avg\\
\hline
LSAN~\cite{cole2021multi} & 70.5 & 87.2 & 52.5 & 18.9 & 57.3\\
ROLE~\cite{cole2021multi} & 70.9 & 89.0 & 50.6 & 20.4 & 57.7\\
LargeLoss~\cite{kim2022large} & 71.6 & 89.3 & 49.6 & 21.8 & 58.1\\
Hill~\cite{zhang2021simple} & 73.2 & 87.8 & 55.0 & 18.8 & 58.7\\
SPLC~\cite{zhang2021simple} & 73.2 & 88.1 & 55.2 & 20.0 & 59.1\\
SCPNet~\cite{ding2023exploring} & 76.4 & 91.2 & 62.0 & 25.7 & 63.8\\
\hline
\textbf{Ours} & \textbf{74.5} & \textbf{89.0} & \textbf{60.0} & \textbf{22.5} & \textbf{61.5}\\
\hline
\end{tabular}
\end{table}

\subsection{Ablation Studies}

We conduct systematic ablation studies to analyze the contribution of each component in our framework and validate key design choices.

\textbf{Component Analysis.} Table~\ref{tab:ablation_components} presents the ablation study on core components. Starting from a Euclidean baseline, we progressively add hyperbolic space embedding with uniformity loss, double-well regularization, and temperature-adaptive classification. Each component contributes to the final performance, with the hyperbolic space foundation providing consistent improvements (+1.6 mAP on COCO), followed by geometric regularization (+0.4 mAP) and the temperature-adaptive mechanism (+0.7 mAP).

\begin{table}[htbp]
\centering
\caption{Ablation study on core components. Results on MS-COCO and PASCAL VOC datasets (mAP \%).}
\label{tab:ablation_components}
\begin{tabular}{|l|c|c|}
\hline
Components & COCO & VOC\\
\hline
Baseline (Euclidean) & 71.8 & 86.5\\
+ Hyperbolic Space + Uniformity Loss & 73.4 & 87.8\\
+ Double-well Regularization & 73.8 & 88.1\\
+ Temperature Adaptive (Full Model) & \textbf{74.5} & \textbf{89.0}\\
\hline
\end{tabular}
\end{table}

\textbf{Temperature Adaptation Analysis.} Table~\ref{tab:temperature_analysis} compares fixed temperature settings with our learnable temperature approach. The adaptive temperature mechanism consistently outperforms fixed alternatives across all datasets, with learned temperatures naturally adapting to different geometric positions in hyperbolic space.

\begin{table}[htbp]
\centering
\caption{Comparison of temperature strategies. Results reported as mAP (\%).}
\label{tab:temperature_analysis}
\begin{tabular}{|l|c|c|c|c|}
\hline
Temperature Strategy & COCO & VOC & NUS-WIDE & CUB\\
\hline
Fixed ($\tau = 0.1$) & 73.2 & 87.8 & 58.5 & 19.9\\
Fixed ($\tau = 0.5$) & 74.0 & 88.4 & 59.2 & 21.1\\
Fixed ($\tau = 1.0$) & 73.8 & 88.1 & 58.9 & 20.8\\
\hline
\textbf{Learnable (Ours)} & \textbf{74.5} & \textbf{89.0} & \textbf{60.0} & \textbf{22.5}\\
\hline
\end{tabular}
\end{table}

\subsection{Analysis and Visualization}
\begin{figure}[htbp]
\centering
\includegraphics[width=0.9\linewidth]{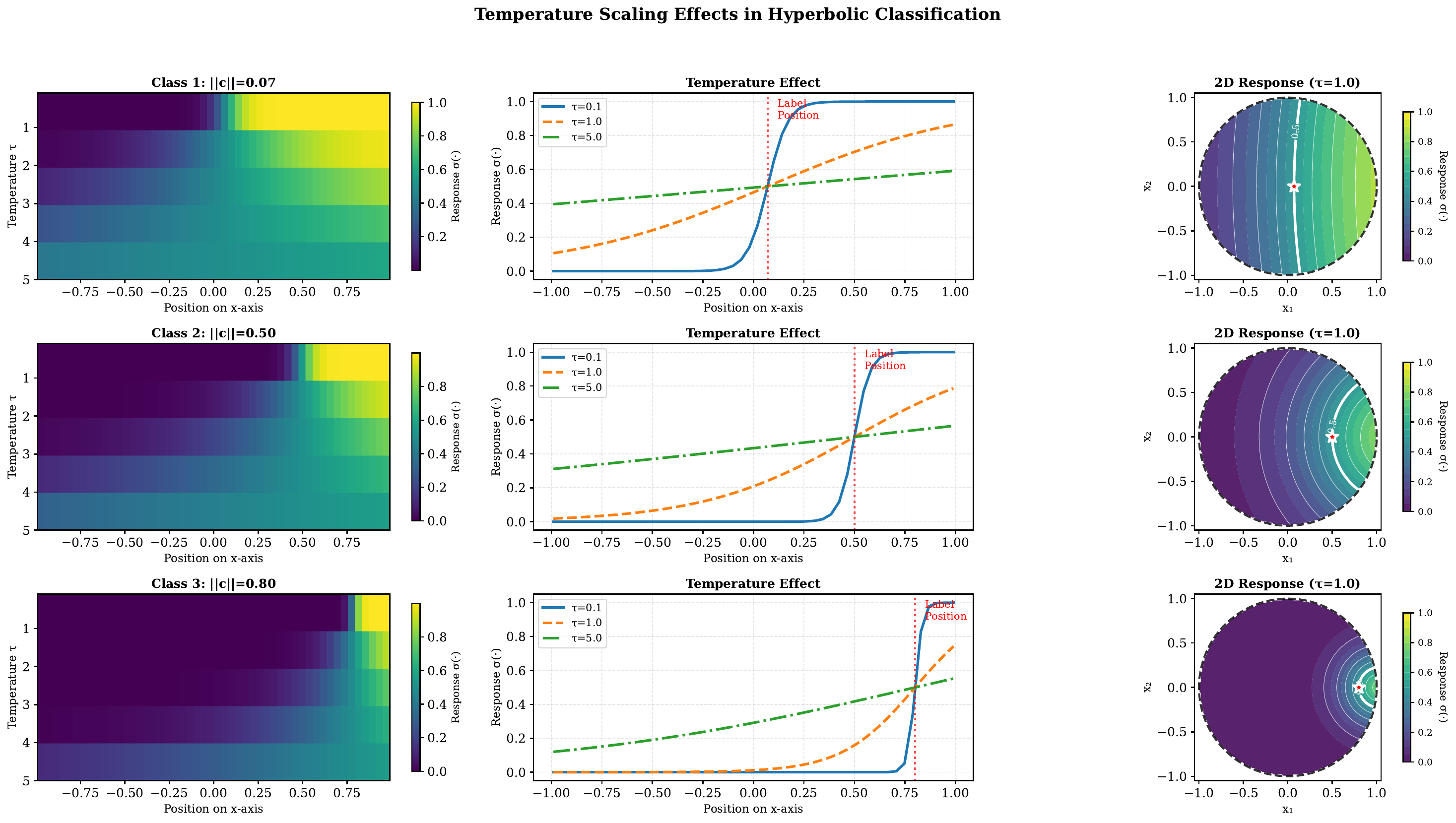}
\caption{Temperature scaling effects in hyperbolic classification. The figure shows comprehensive analysis across three class positions ($\|c\| = 0.07, 0.5, 0.8$): (left column) temperature-coordinate heatmaps, (middle column) response curves for different temperatures, (right column) 2D response maps with $\tau = 1.0$. Temperature scaling enables adaptive decision boundaries that respect hyperbolic geometry properties.}
\label{fig:temperature_analysis_2}
\end{figure}

\textbf{Comprehensive Temperature Analysis.} Figures~\ref{fig:temperature_analysis} and \ref{fig:temperature_analysis_2} provide comprehensive visualization of our temperature scaling approach across different geometric positions in hyperbolic space. The analysis combines 2D spatial visualization and 1D response curve analysis to demonstrate systematic advantages of temperature adaptation. The 2D response maps reveal how temperature scaling creates position-aware classification boundaries that adapt to hyperbolic geometry: low temperatures ($\tau = 0.1$) produce sharp boundaries but may cause overconfident predictions, while high temperatures ($\tau = 5.0$) generate smooth surfaces but sacrifice discriminative power. Our adaptive mechanism learns to balance these trade-offs optimally. The 1D response curve analysis demonstrates consistent superiority across all positions, with the most pronounced improvements for boundary-proximate labels where compressed decision regions require careful calibration.

\begin{figure}[htbp]
\centering
\includegraphics[width=0.75\linewidth]{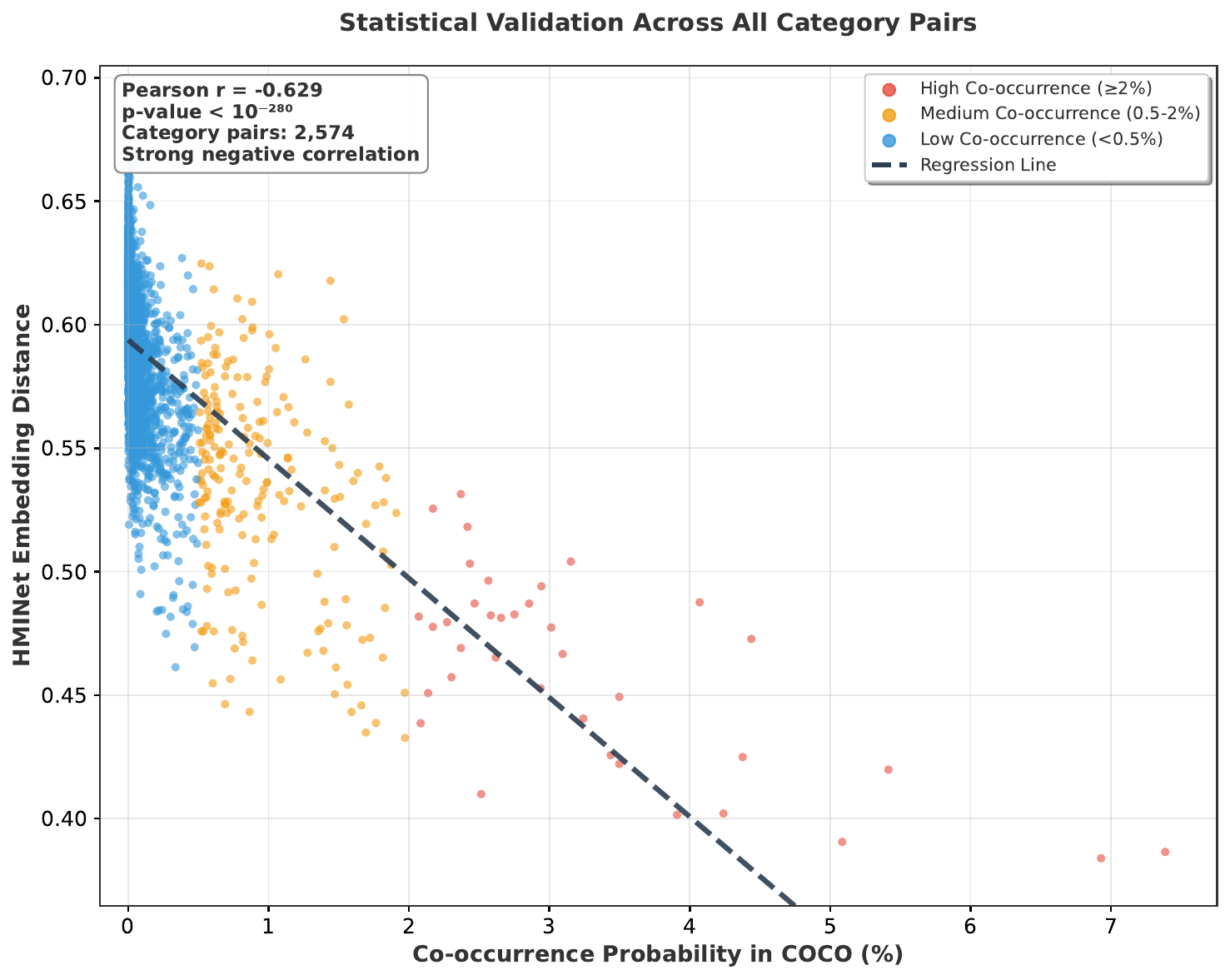}
\caption{Statistical validation of co-occurrence pattern learning in COCO dataset. The scatter plot shows the relationship between real-world co-occurrence probability and learned hyperbolic embedding center distances across 2,574 category pairs. Strong negative correlation (Pearson $r = -0.629$, $p < 10^{-280}$) demonstrates that our model automatically captures meaningful statistical patterns from the data. Color coding distinguishes frequency groups: red (high co-occurrence $\geq 2\%$), orange (medium $0.5$--$2\%$), and blue (low $< 0.5\%$). The systematic organization validates that our ball-based framework learns meaningful label relationships rather than arbitrary spatial arrangements.}
\label{fig:cooccurrence_analysis}
\end{figure}

\textbf{Co-occurrence Pattern Learning Analysis.} Figure~\ref{fig:cooccurrence_analysis} validates that our ball-based framework learns interpretable relationships by analyzing COCO co-occurrence statistics versus learned hyperbolic embedding center distances. We observe strong negative correlation (Pearson $r = -0.629$, $p < 10^{-280}$) across 2,574 category pairs, demonstrating our framework automatically captures real-world statistical patterns without explicit supervision, providing transparent insights into discovered semantic relationships.

These comprehensive experiments demonstrate that our hyperbolic structured classification framework effectively leverages the geometric properties of hyperbolic space to learn meaningful label relationships under incomplete supervision. The consistent improvements across diverse datasets and the interpretable geometric organization of learned representations validate our approach's effectiveness for single positive multi-label learning.

\section{Conclusion}

This work introduces hyperbolic ball-based classification, the first framework to represent labels as interacting geometric balls for single positive multi-label learning. Unlike traditional SPMLL methods that use point representations in Euclidean space, our approach leverages hyperbolic geometry to naturally capture hierarchical inclusion, co-occurrence overlap, and semantic separation through ball-to-ball interactions.

Our framework combines learnable hyperbolic ball representations with temperature-adaptive classification and physics-inspired double-well regularization, enabling rich inter-label relationships to emerge automatically during training. Extensive experiments on four benchmark datasets show that our method achieves competitive performance while offering superior interpretability. Statistical analysis further confirms that the learned embeddings correlate strongly with real-world co-occurrence patterns, indicating that the geometric organization captures meaningful semantic relationships.

This work establishes hyperbolic geometry as a powerful paradigm for structured classification under incomplete supervision, with broad implications for hierarchical classification and geometric machine learning. By demonstrating that complex semantic relationships can emerge naturally from geometric constraints, our approach opens new avenues for interpretable learning from minimal supervision.



{
    \small
    \bibliographystyle{IEEEtran}

\begin{thebibliography}{10}
\providecommand{\url}[1]{#1}
\csname url@samestyle\endcsname
\providecommand{\newblock}{\relax}
\providecommand{\bibinfo}[2]{#2}
\providecommand{\BIBentrySTDinterwordspacing}{\spaceskip=0pt\relax}
\providecommand{\BIBentryALTinterwordstretchfactor}{4}
\providecommand{\BIBentryALTinterwordspacing}{\spaceskip=\fontdimen2\font plus
\BIBentryALTinterwordstretchfactor\fontdimen3\font minus \fontdimen4\font\relax}
\providecommand{\BIBforeignlanguage}[2]{{%
\expandafter\ifx\csname l@#1\endcsname\relax
\typeout{** WARNING: IEEEtran.bst: No hyphenation pattern has been}%
\typeout{** loaded for the language `#1'. Using the pattern for}%
\typeout{** the default language instead.}%
\else
\language=\csname l@#1\endcsname
\fi
#2}}
\providecommand{\BIBdecl}{\relax}
\BIBdecl

\bibitem{lv2019weakly}
J.~Lv, N.~Xu, R.~Zheng, and X.~Geng, ``Weakly supervised multi-label learning via label enhancement.'' in \emph{IJCAI}, 2019, pp. 3101--3107.

\bibitem{cole2021multi}
E.~Cole, O.~Mac~Aodha, T.~Lorieul, P.~Perona, D.~Morris, and N.~Jojic, ``Multi-label learning from single positive labels,'' in \emph{Proceedings of the IEEE/CVF Conference on Computer Vision and Pattern Recognition}, 2021, pp. 933--942.

\bibitem{zhang2021simple}
Y.~Zhang, Y.~Cheng, X.~Huang, F.~Wen, R.~Feng, Y.~Li, and Y.~Guo, ``Simple and robust loss design for multi-label learning with missing labels,'' \emph{arXiv preprint arXiv:2112.07368}, 2021.

\bibitem{kim2022large}
Y.~Kim, J.~M. Kim, Z.~Akata, and J.~Lee, ``Large loss matters in weakly supervised multi-label classification,'' in \emph{Proceedings of the IEEE/CVF Conference on Computer Vision and Pattern Recognition}, 2022, pp. 14\,156--14\,165.

\bibitem{kim2023bridging}
Y.~Kim, J.~M. Kim, J.~Jeong, C.~Schmid, Z.~Akata, and J.~Lee, ``Bridging the gap between model explanations in partially annotated multi-label classification,'' in \emph{Proceedings of the IEEE/CVF Conference on Computer Vision and Pattern Recognition}, 2023, pp. 3408--3417.

\bibitem{ding2023exploring}
Z.~Ding, A.~Wang, H.~Chen, Q.~Zhang, P.~Liu, Y.~Bao, W.~Yan, and J.~Han, ``Exploring structured semantic prior for multi label recognition with incomplete labels,'' in \emph{Proceedings of the IEEE/CVF Conference on Computer Vision and Pattern Recognition}, 2023, pp. 3398--3407.

\bibitem{liu2023revisiting}
B.~Liu, N.~Xu, J.~Lv, and X.~Geng, ``Revisiting pseudo-label for single-positive multi-label learning,'' in \emph{International Conference on Machine Learning}.\hskip 1em plus 0.5em minus 0.4em\relax PMLR, 2023, pp. 22\,249--22\,265.

\bibitem{wang2023hierarchical}
A.~Wang, H.~Chen, Z.~Lin, Z.~Ding, P.~Liu, Y.~Bao, W.~Yan, and G.~Ding, ``Hierarchical prompt learning using clip for multi-label classification with single positive labels,'' in \emph{Proceedings of the 31st ACM International Conference on Multimedia}, 2023, pp. 5594--5604.

\bibitem{NEURIPS2022_d51ab0fc}
B.~Xiong, M.~Cochez, M.~Nayyeri, and S.~Staab, ``Hyperbolic embedding inference for structured multi-label prediction,'' in \emph{Advances in Neural Information Processing Systems}, S.~Koyejo, S.~Mohamed, A.~Agarwal, D.~Belgrave, K.~Cho, and A.~Oh, Eds., vol.~35.\hskip 1em plus 0.5em minus 0.4em\relax Curran Associates, Inc., 2022, pp. 33\,016--33\,028.

\bibitem{lin2014microsoft}
T.-Y. Lin, M.~Maire, S.~Belongie, J.~Hays, P.~Perona, D.~Ramanan, P.~Doll{\'a}r, and C.~L. Zitnick, ``Microsoft coco: Common objects in context,'' in \emph{European conference on computer vision}.\hskip 1em plus 0.5em minus 0.4em\relax Springer, 2014, pp. 740--755.

\bibitem{hoiem2009pascal}
D.~Hoiem, S.~K. Divvala, and J.~H. Hays, ``Pascal voc 2008 challenge,'' \emph{World Literature Today}, vol.~24, no.~1, pp. 1--4, 2009.

\bibitem{chua2009nus}
T.-S. Chua, J.~Tang, R.~Hong, H.~Li, Z.~Luo, and Y.~Zheng, ``Nus-wide: a real-world web image database from national university of singapore,'' in \emph{Proceedings of the ACM international conference on image and video retrieval}, 2009, pp. 1--9.

\bibitem{wah2011caltech}
C.~Wah, S.~Branson, P.~Welinder, P.~Perona, and S.~Belongie, ``The caltech-ucsd birds-200-2011 dataset,'' 2011.

\bibitem{chen2020hyperbolic}
B.~Chen, X.~Huang, L.~Xiao, Z.~Cai, and L.~Jing, ``Hyperbolic interaction model for hierarchical multi-label classification,'' in \emph{Proceedings of the AAAI conference on artificial intelligence}, vol.~34, no.~05, 2020, pp. 7496--7503.

\bibitem{chatterjee2021joint}
S.~Chatterjee, A.~Maheshwari, G.~Ramakrishnan, and S.~N. Jagaralpudi, ``Joint learning of hyperbolic label embeddings for hierarchical multi-label classification,'' \emph{arXiv preprint arXiv:2101.04997}, 2021.

\bibitem{nickel2018learning}
M.~Nickel and D.~Kiela, ``Learning continuous hierarchies in the lorentz model of hyperbolic geometry,'' in \emph{International conference on machine learning}.\hskip 1em plus 0.5em minus 0.4em\relax PMLR, 2018, pp. 3779--3788.

\bibitem{ganea2018hyperbolic}
O.~Ganea, G.~B{\'e}cigneul, and T.~Hofmann, ``Hyperbolic entailment cones for learning hierarchical embeddings,'' in \emph{International conference on machine learning}.\hskip 1em plus 0.5em minus 0.4em\relax PMLR, 2018, pp. 1646--1655.

\bibitem{chen2023label}
C.-Y. Chen, T.-M. Hung, Y.-L. Hsu, and L.-W. Ku, ``Label-aware hyperbolic embeddings for fine-grained emotion classification,'' \emph{arXiv preprint arXiv:2306.14822}, 2023.

\bibitem{tifrea2018poincar}
A.~Tifrea, G.~B{\'e}cigneul, and O.-E. Ganea, ``Poincar$\backslash$'e glove: Hyperbolic word embeddings,'' \emph{arXiv preprint arXiv:1810.06546}, 2018.

\bibitem{cho2019large}
H.~Cho, B.~DeMeo, J.~Peng, and B.~Berger, ``Large-margin classification in hyperbolic space,'' in \emph{The 22nd international conference on artificial intelligence and statistics}.\hskip 1em plus 0.5em minus 0.4em\relax PMLR, 2019, pp. 1832--1840.

\bibitem{zhou2022acknowledging}
D.~Zhou, P.~Chen, Q.~Wang, G.~Chen, and P.-A. Heng, ``Acknowledging the unknown for multi-label learning with single positive labels,'' in \emph{European Conference on Computer Vision}.\hskip 1em plus 0.5em minus 0.4em\relax Springer, 2022, pp. 423--440.

\bibitem{xing2024vision}
X.~Xing, Z.~Xiong, A.~Stylianou, S.~Sastry, L.~Gong, and N.~Jacobs, ``Vision-language pseudo-labels for single-positive multi-label learning,'' in \emph{Proceedings of the IEEE/CVF Conference on Computer Vision and Pattern Recognition}, 2024, pp. 7799--7808.

\bibitem{imambi2021pytorch}
S.~Imambi, K.~B. Prakash, and G.~Kanagachidambaresan, ``Pytorch,'' in \emph{Programming with TensorFlow: solution for edge computing applications}.\hskip 1em plus 0.5em minus 0.4em\relax Springer, 2021, pp. 87--104.

\bibitem{kochurov2020geoopt}
M.~Kochurov, R.~Karimov, and S.~Kozlukov, ``Geoopt: Riemannian optimization in pytorch,'' \emph{arXiv preprint arXiv:2005.02819}, 2020.

\bibitem{zhang2025semantic}
R.~Zhang, H.~Qiao, P.~Xu, M.~Shang, and L.~Chen, ``Semantic-guided representation learning for multi-label recognition,'' \emph{arXiv preprint arXiv:2504.03801}, 2025.

\bibitem{wang2025splicemix}
L.~Wang, Y.~Zhan, L.~Ma, D.~Tao, L.~Ding, and C.~Gong, ``Splicemix: A cross-scale and semantic blending augmentation strategy for multi-label image classification,'' \emph{IEEE Transactions on Multimedia}, 2025.

\bibitem{chen2024boosting}
Y.~Chen, C.~Li, X.~Dai, J.~Li, W.~Sun, Y.~Wang, R.~Zhang, T.~Zhang, and B.~Wang, ``Boosting single positive multi-label classification with generalized robust loss,'' \emph{arXiv preprint arXiv:2405.03501}, 2024.

\bibitem{xu2022one}
N.~Xu, C.~Qiao, J.~Lv, X.~Geng, and M.-L. Zhang, ``One positive label is sufficient: Single-positive multi-label learning with label enhancement,'' \emph{Advances in Neural Information Processing Systems}, vol.~35, pp. 21\,765--21\,776, 2022.

\bibitem{hu2025co}
T.~Hu, W.~Zhang, J.~Guo, and H.~Li, ``Co-pseudo labeling and active selection for fundus single-positive multi-label learning,'' \emph{IEEE Transactions on Medical Imaging}, 2025.

\bibitem{song2024ignore}
G.~Song, N.-r. Kim, J.-S. Lee, and J.-H. Lee, ``Ignore: Information gap-based false negative loss rejection for single positive multi-label learning,'' in \emph{European Conference on Computer Vision}.\hskip 1em plus 0.5em minus 0.4em\relax Springer, 2024, pp. 472--488.

\bibitem{yiming2023context}
Y.~Lin, X.-B. Jin, Q.~Wang, and K.~Huang, ``Context does matter: End-to-end panoptic narrative grounding with deformable attention refined matching network,'' in \emph{2023 IEEE International Conference on Data Mining (ICDM)}, 2023, pp. 1163--1168.

\end{thebibliography}

}

\end{document}